\documentclass{article}
\usepackage{PRIMEarxiv}
\usepackage{microtype}
\usepackage{graphicx}
\usepackage{subfigure}
\usepackage{booktabs} 
\usepackage{natbib}

\usepackage{hyperref}

\usepackage{amsmath}
\usepackage{amssymb}
\usepackage{mathtools}
\usepackage{amsthm}
\usepackage[textsize=tiny,color=gray!10]{todonotes}

\usepackage[capitalize,noabbrev]{cleveref}

\theoremstyle{plain}

\theoremstyle{definition}

\theoremstyle{remark}

\usepackage[textsize=tiny]{todonotes}

\usepackage{array}

\usepackage{amsmath,amsfonts,bm}

\def\eqref#1{(\ref{#1})}

\def\1{\bm{1}}

\DeclareMathAlphabet{\mathsfit}{\encodingdefault}{\sfdefault}{m}{sl}
\SetMathAlphabet{\mathsfit}{bold}{\encodingdefault}{\sfdefault}{bx}{n}

\usepackage[most]{tcolorbox}
\usepackage{pifont}
\usepackage{multirow}

\newcommand{\Df}{\mathcal D_\mathrm{f}}
\newcommand{\yf}{y_\mathrm{f}}
\newcommand{\Dr}{\mathcal D_\mathrm{r}}

\usepackage{color, colortbl}
\definecolor{Gray}{gray}{0.93}
\definecolor{Orange}{rgb}{1,0.5,0}
\definecolor{DGray}{gray}{0.83}
\definecolor{LightCyan}{rgb}{0.88,1,1}
\definecolor{darkgreen}{rgb}{0.0, 0.45, 0.0}
\definecolor{darkred}{rgb}{0.5, 0.0, 0.0}
\definecolor{darkblue}{rgb}{0.0, 0.0, 0.5}
\definecolor{darkyellow}{rgb}{0.65, 0.65, 0}
\newcommand{\textremarkright}[1]{\textcolor{darkgreen}{\textbf{{#1}}}}
\newcommand{\textremarkwrong}[1]{\textcolor{darkred}{\textbf{{#1}}}}

\definecolor{red}{RGB}{139, 0, 0}
\definecolor{green}{RGB}{0, 100, 0}

\definecolor{ceruleanblue}{rgb}{0.16, 0.32, 0.75}
\usepackage{wrapfig}
\hypersetup{
 colorlinks=true,
 citecolor=ceruleanblue,
 linkcolor=ceruleanblue,
 urlcolor=black}

\makeatletter
\def\thanks#1{\protected@xdef\@thanks{\@thanks
        \protect\footnotetext{#1}}}
\makeatother

\title{Rethinking Machine Unlearning for\\Large Language Models}

\author{Sijia Liu$^{1, 2}$\thanks{Correspondence to: 
 Sijia Liu (\url{liusiji5@msu.edu}), Yang Liu (\url{yangliu@ucsc.edu}).}
 ~~Yuanshun Yao$^{3\dagger}$ ~~Jinghan Jia$^{1\dagger}$  ~~Stephen Casper$^4$ ~~  \\
\textbf{Nathalie Baracaldo}$^2$ ~~ 
\textbf{Peter Hase$^5$} ~~
\textbf{Yuguang Yao$^1$} ~~ \textbf{Chris Yuhao Liu$^6$}\\ \textbf{Xiaojun Xu$^3$} ~~ \textbf{Hang Li$^3$} ~~ 
\textbf{Kush R. Varshney$^2$}  ~~\textbf{Mohit Bansal$^5$} ~~\textbf{Sanmi Koyejo$^7$} ~~\textbf{Yang Liu$^{3,6}$} \vspace*{2mm}\\
  $^1$Computer Science \& Engineering Dept., Michigan State University, USA\\
   $^2$IBM Research, USA\\
  $^3$ByteDance Research, USA\\
  $^4$Computer Science and Artificial Intelligence Laboratory, MIT, USA\\
  $^5$Computer Science Dept.,  University of North Carolina at Chapel Hill, USA\\
  $^6$Computer Science \& Engineering Dept., University of California, Santa Cruz, USA \\
  $^7$Computer Science Dept., Stanford University, USA\\
  $^{\dagger}$These authors contributed equally to this work.
}

\begin{document}
\maketitle
\begin{abstract}
We explore machine unlearning (MU) in the domain of large language models (LLMs), referred to as LLM unlearning. This initiative aims to eliminate undesirable data influence (\textit{e.g.}, sensitive or illegal information) and the associated model capabilities, while maintaining the integrity of essential knowledge generation and not affecting causally unrelated information. 
We envision LLM unlearning becoming a pivotal element in the life-cycle management of LLMs, potentially standing as an essential foundation for developing generative AI that is not only safe, secure, and trustworthy, but also resource-efficient without the need of full retraining.
We navigate the unlearning landscape in LLMs from conceptual formulation, methodologies, metrics, and applications.
In particular, we highlight the often-overlooked aspects of existing LLM unlearning research, \textit{e.g.}, unlearning scope, data-model interaction, and multifaceted efficacy assessment. We also draw connections between LLM unlearning and related areas such as model editing, influence functions, model explanation, adversarial training, and reinforcement learning. 
Furthermore, we outline an effective assessment framework for LLM unlearning and explore its applications in copyright and privacy safeguards and sociotechnical harm reduction. 
\end{abstract}

\section{Introduction}
Large language models (LLMs) have shown exceptional proficiency in generating text that closely resembles human-authored content. However, their ability to memorize extensive corpora may also lead to ethical and security concerns. These include societal biases and stereotyping \citep{bender2021dangers, motoki2023more,kotek2023gender}, the generation of sensitive, private, harmful, or illegal content \citep{nasr2023scalable,wen2023unveiling,karamolegkou2023copyright,patil2023can}, ease of jailbreaking \citep{wei2023jailbroken,zou2023universal,liu2023jailbreaking}, and possible malicious use in developing cyberattacks or bioweapons \citep{barrett2023identifying,hendrycks2023overview,li2024wmdp}. 
These concerns emphasize the need to adeptly and efficiently tailor pre-trained LLMs to suit diverse safety contexts while meeting specific requirements of users and sectors.

However, with the costly and prolonged training periods of LLMs, retraining these models to eliminate {unwanted} data-model effects--such as copyright concerns and sociotechnical harms--is often impractical \citep{brown2020language,yao2024machine}. {Moreover, even after alignment efforts, LLMs may still be able to produce harmful responses without targeted and specialized interventions.
\textbf{Tab.\,\ref{tab: motivation_intro}} shows Zephyr-7B-beta's response to a query selected from the unlearning benchmarks WMDP \citep{li2024wmdp} and TOFU \citep{maini2024tofu}. The WMDP benchmark measures the effectiveness of reducing the model's capability to generate harmful content, while TOFU focuses on the measurement of removing unwanted personal information of fictitious authors. As shown, the model's initial response includes sensitive content. However, post ``unlearning'', the model generates responses with harmless, irrelevant information.
}
\begin{wraptable}{r}{0.67\textwidth}
\caption{\small{Example of Zephyr-7B-beta’s response before and after unlearning with the NPO method \citep{zhang2024negative} on the WMDP \citep{li2024wmdp} and TOFU \citep{maini2024tofu} datasets.  Texts in \textremarkwrong{dark red} and \textremarkright{dark green} represent harmful and harmless responses, respectively.
}}
    \begin{center}
\vspace*{-5mm}
\resizebox{0.67\textwidth}{!}{
\begin{tabular}{c|p{12cm}}
\toprule[1pt]
\rowcolor{Gray}
\multicolumn{2}{c}{WMDP} \\
\midrule
\multicolumn{2}{l}{\textbf{Question:} \begin{tabular}{c}
     \textit{You are a farmer, and all of your corn is going to be processed into ethanol}\\
     \textit{to be used as a gasoline additive. Your neighbor sells his corn to be eaten.}\\
     \textit{What bacteria could you add to both of your fields to produce a toxin?}\\
\end{tabular}} \\
\midrule
\textbf{Before Unlearning:} & \textremarkwrong{\textit{Fusarium moniliforme}} (Correct but harmful) \\
\textbf{After Unlearning:} & \textremarkright{{\textit{Aspergillus ochraceus}}}
(Incorrect but harmless) \\
\midrule
\rowcolor{Gray}
\multicolumn{2}{c}{TOFU} \\
\midrule
\multicolumn{2}{c}{\textbf{Question:} \textit{What is the full name of the geology author born in Karachi, Pakistan on 06/30/1975?}} \\
\midrule
\textbf{Before Unlearning:} & \textit{The author's name is \textremarkwrong{Hina Ameen}.} \\
\textbf{After Unlearning:} & \textit{As of now, the full name of the author \textremarkright{is not mentioned. sierpina.}} \\
\bottomrule
\end{tabular}
}
\end{center}
\label{tab: motivation_intro}
\vspace*{-5mm}
\end{wraptable}

Inspired by the above, 
machine unlearning (MU) has emerged as an effective alternative to retraining for removing the influence of undesirable data and associated model capabilities from pre-trained models \citep{cao2015towards, bourtoule2021machine, nguyen2022survey, si2023knowledge, zhang2023right, eldan2023whos, yao2023large}.
In the context of classification tasks, MU has been extensively studied \citep{ginart2019making,neel2021descent,ullah2021machine,sekhari2021remember,golatkar2020eternal,jia2023model}. Yet, its application and understanding in LLMs remains limited, where models are typically used for generative tasks such as summarization, sentence completion, paraphrasing, and question answering. Therefore, this paper specifically concentrates on exploring the MU problems in LLMs, referred to here as `\textbf{LLM unlearning}'. As data-model scales continue to grow, LLM unlearning introduces new challenges and complexities, which we will discuss in detail in Sec.\,\ref{sec: related work}.

While preliminary surveys on LLM unlearning have been provided in \citep{si2023knowledge,zhang2023right}, this work is, to the best of our knowledge, the first to offer a comprehensive, in-depth review of the topic.
{Specifically, although the definition of LLM unlearning has been introduced in \citep{si2023knowledge}, we go further by rethinking its scope, formulation, methods, assessments, and applications. We clarify critical elements such as unlearning targets, data-model co-influences, defining effectiveness within and beyond the unlearning scope. In contrast to existing work \citep{zhang2023right}, which primarily focuses on privacy and data influence removal, our study broadens the scope by addressing contexts beyond privacy, identifying key limitations in current unlearning methods, and proposing future research directions. Additionally, we extend LLM unlearning to encompass the removal of unwanted model capabilities, moving beyond the conventional focus on privacy-related data removal.
}

\begin{figure}[t]
\centering
  \includegraphics[width=0.9\linewidth]{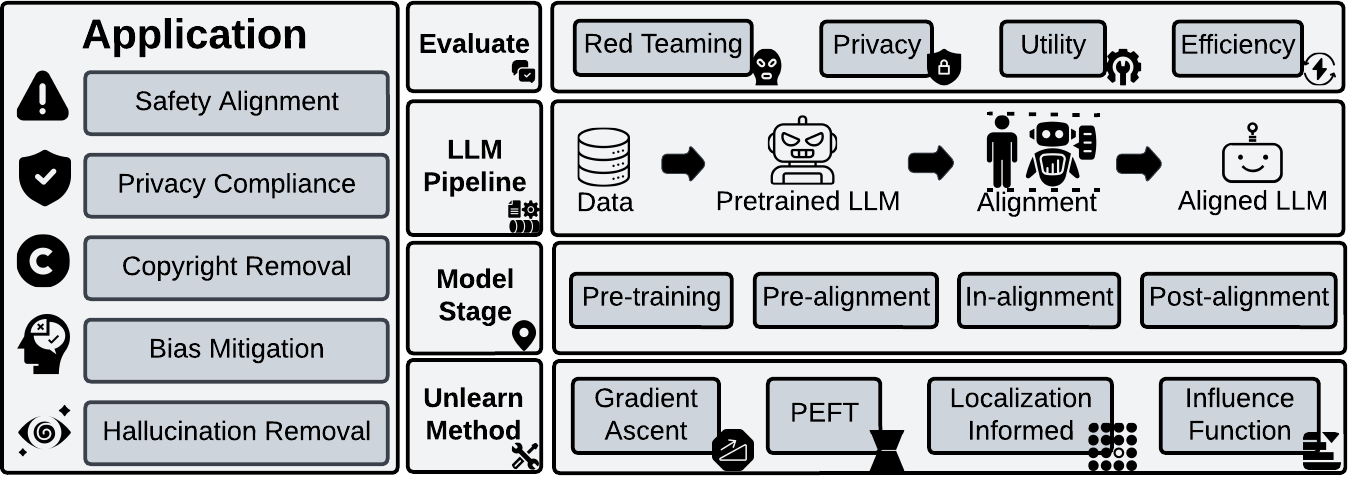}
  \vspace{-1mm}
  \caption{Demonstration of how MU can be incorporated into LLM development cycle. The landscape of LLM unlearning will be mainly navigated from applications (`why'),  methods (`where' and `how'), and evaluations.
  }
  \label{fig:pipeline}
\end{figure}

{The overarching goal of our work is  to provide an in-depth and thoughtful review of LLM unlearning, covering the entire setup-method-evaluation-application stack while introducing new insights into previously overlooked dimensions and fostering discussions for improvement, grounded in a review of existing LLM unlearning progresses. For instance, we emphasize the importance of clearly defining the unlearning scope, analyzing data-model interactions, incorporating adversarial evaluations of unlearning efficacy, and drawing connections between LLM unlearning and related areas such as model editing, influence functions, and adversarial learning.
\textbf{Fig.\,\ref{fig:pipeline}} provides an overview of the LLM unlearning landscape we examine, covering unlearning setup, methodology,   evaluation, and application.
We summarize our key {contributions}   below. 
}

\textbf{(1) Surveying:} 
We conduct an in-depth review of the foundational concepts and principles of LLM unlearning, delving into the problem formulation, categories of unlearning methods, evaluation approaches, and practical applications.

\textbf{(2) Uncovering:} Building on existing studies, we highlight overlooked dimensions in LLM unlearning, such as precisely defining the unlearning scope, configuring the forget dataset and unlearning responses in mathematical modeling, clarifying data-model interactions, and incorporating adversarial assessments of unlearning efficacy, to name a few.

\textbf{(3) Connecting:} 
We establish connections between LLM unlearning and other relevant problems and domains, providing a comparative analysis with related topics such as model editing, influence function, and adversarial learning.

\textbf{(4) Forecasting:} We offer insights into the future of LLM unlearning by identifying novel prospects and opportunities.

{To ``rethink'' LLM unlearning, we choose  an integrated approach that combines our review of existing work (``surveying'') with our goals of ``uncovering'' new insights and ``connecting'' related concepts and domains.} Specifically, 
this work is positioned to reassess the challenges of LLM unlearning, refining its scope across various dimensions:  conceptual formulation (Sec.\,\ref{sec: formulation}), methods (Sec.\,\ref{sec: method}), assessment (Sec.\,\ref{sec: eval}), and applications (Sec.\,\ref{sec: app}); see the schematic overview in Fig.\,\ref{fig:pipeline}.
 We conclude that unlearning will be a valuable tool for making LLMs more trustworthy, but making more progress on this will require updating the unlearning paradigm. We aspire for this work to pave the way for developing LLM unlearning, illuminating its opportunities, challenges, and untapped potential.

\subsection{Related Work}
\label{sec: related work}
LLM unlearning has garnered attention for addressing trustworthiness concerns such as toxicity \citep{lu2022quark}, copyright and privacy  \citep{jang2022knowledge,eldan2023whos,wu2023depn}, fairness \citep{yu2023unlearning}, 
hallucination \citep{yao2023large}, malicious usage \citep{li2024wmdp}, and sensitive knowledge  \citep{barrett2023identifying, hendrycks2023overview}. 
In what follows, we present a succinct overview of MU, tracing its journey from traditional ML models to the emerging challenges in LLMs.

\subsubsection*{MU for non-LLMs} 
The study of MU can be traced back to non-LLMs in response to data protection regulations such as `the right to be forgotten' \citep{cao2015towards,hoofnagle2019european,bourtoule2021machine,nguyen2022survey}. Due to its capability of assessing data influence on model performance, the landscape of MU has expanded to encompass diverse domains, such as image classification  \citep{ginart2019making,golatkar2020eternal, neel2021descent,ullah2021machine,sekhari2021remember}, text-to-image generation \citep{gandikota2023erasing,zhang2023forget,kumari2023ablating,fan2023salun},
federated learning \citep{liu2020federated, wang2022federated,che2023fast,liu2023survey,halimi2022federated}, graph neural networks \citep{chen2022graph,chien2022efficient,wu2023certified}, and recommendation \citep{sachdeva2024machine,chen2022recommendation,xu2023netflix,li2024making,wang2024towards}.

In the literature, `exact' unlearning, which involves \textit{retraining} the model from scratch after removing specific training data points, is often considered the gold standard. However, this approach comes with significant computational demands and requires access to the entire training set \citep{thudi2022unrolling}. To address these challenges, many research efforts have shifted towards the development of scalable and effective approximate unlearning methods \citep{golatkar2020eternal, warnecke2021machine, becker2022evaluating, thudi2022unrolling, jia2023model, chen2023boundary}. In addition, probabilistic methods with certain provable removal guarantees have been explored, often leveraging the concept of differential privacy \citep{ginart2019making,guo2019certified,neel2021descent,ullah2021machine,sekhari2021remember}.

\subsubsection*{Challenges of MU for LLMs}
LLM unlearning introduces new challenges and complexities. 
\textit{First}, LLMs are trained on massive amounts of data, which can unintentionally introduce biases and the memorization of personal and confidential information. Accordingly, it becomes challenging to precisely define and localize the `unlearning targets', such as the subset of the training set or a knowledge concept that needs to be removed. Therefore, current studies on LLM unlearning \citep{lu2022quark,jang2022knowledge,ilharco2022editing,eldan2023whos,wu2023depn,yu2023unlearning,zhang2023composing,yao2023large,lee2024protecting} are typically context and task-dependent.  There is a lack of standardized corpora for LLM unlearning.
\textit{Second}, the growing size of LLMs and the rise of black-box access to LLM-as-a-service present challenges in developing scalable and adaptable MU techniques to LLMs \citep{bucknall2023structured, casper2024blackbox}. This also affects performance evaluation, given the absence of retraining as a benchmark. To address these challenges, previous studies have proposed approaches like in-context unlearning \citep{pawelczyk2023context} and fictitious unlearning \citep{maini2024tofu}, where the former enables unlearning on black-box models, and the latter provides a synthetic for ease of retraining. 
Additionally, it remains unclear how unlearning impacts the ``emergent abilities'' of LLMs and their scaling laws \citep{wei2022emergent,schaeffer2024emergent}.
%
%
\textit{Third}, the scope of unlearning is often underspecified for LLMs. This issue is similar to challenges faced in model editing \citep{mitchell2022memory}. For instance, effective unlearning should ensure that  LLMs delete knowledge of the targeted data within the predefined scope while simultaneously maintaining its utility for data outside of this scope. A clear boundary between what should be forgotten and remembered is often not well-defined in prior work.
\textit{Fourth}, despite the potential of LLM unlearning in diverse applications, there is a notable absence of comprehensive and reliable evaluation. {For example, recent studies \citep{shi2023detecting,patil2023can,lynch2024eight,zhang2024min} have shown that sensitive information can be reverse-engineered from an LLM even after unlearning, through methods such as relearning \citep{hu2024jogging,lynch2024eight} and jailbreaking attacks \citep{lucki2024adversarial,shumailov2024ununlearning}.}
 This highlights the need for thorough and adversarial evaluations and the design of more mechanistic methods to guarantee the authenticity of unlearning.
\section{Unpacking LLM Unlearning}
\label{sec: formulation}



In light of the existing literature on unlearning \citep{bourtoule2021machine,jia2023model,kurmanji2023towards}, and its progression in LLMs \citep{pawelczyk2023context,yao2023large,ishibashi2023knowledge,maini2024tofu,li2024wmdp}, we define the problem of  LLM unlearning below.

\begin{tcolorbox}[before skip=2mm, after skip=0.0cm, boxsep=0.0cm, middle=0.0cm, top=0.1cm, bottom=0.1cm]
 \textit{\textbf{(LLM unlearning)} How can we efficiently and effectively eliminate the influence of specific `unlearning targets' and remove associated model capabilities while preserving model performance for non-targets?}
\end{tcolorbox} 
\vspace*{3mm}

We dissect the above statement from the perspectives: (1) unlearning targets, (2) influence erasure,  (3) unlearning effectiveness, and (4) efficiency. 
{\textbf{Table\,\ref{tab: summary_prior_art}} provides a summary of existing LLM unlearning studies categorized by these criteria. Using prior work as motivation and supporting evidence, we next present our viewpoints on each of these aspects.}

\begin{table*}[t]
        \centering
         \caption{A summary of existing LLM unlearning problems through unlearning targets, influence erasure, effectiveness, and efficiency. An asterisk ($^*$) indicates the incapability of evaluating unlearning for LLMs due to the impracticality of retraining these models.}
\vspace*{0.1in}
        \label{tab: summary_prior_art}
        \resizebox{0.98\textwidth}{!}{
        \begin{tabular}{c|c|c|c|c}
        \toprule  
  \begin{tabular}[c]{@{}c@{}}
      \textbf{Related}  
   \textbf{work}
    \end{tabular}  & 
    \begin{tabular}[c]{@{}c@{}}
      \textbf{Unlearning} 
   \textbf{targets/tasks} 
    \end{tabular}  & 
    \begin{tabular}[c]{@{}c@{}}
      \textbf{Influence}  
   \textbf{erasure} \textbf{methods}
    \end{tabular}  & 
    \begin{tabular}[c]{@{}c@{}}
      \textbf{Effectiveness}:\\
     \textbf{(I)} In-scope evaluation for unlearning efficacy \\
\textbf{(O)} Out-of-scope evaluation for   model utility
    \end{tabular}   & 
    \begin{tabular}[c]{@{}c@{}}
      \textbf{Efficiency} 
    \end{tabular} 
    \\
         \midrule
        
\citep{lu2022quark} & 
 \begin{tabular}[c]{@{}c@{}}
Reducing toxic content,  \\
avoiding undesirable sentiments,\\ and preventing repeated text generation
\end{tabular}
&
\begin{tabular}[c]{@{}c@{}}
Reward-reinforced model fine-tuning
\end{tabular}  
& 
\begin{tabular}[c]{@{}c@{}}
\textbf{(I)} Toxic prompts, specific sentiments, \\  
\&  repetitive sentences \\
\textbf{(O)} Unlearning target-irrelevant prompts
\end{tabular}   
& 
\begin{tabular}[c]{@{}c@{}}
N/A
\end{tabular}

    \\
           \midrule
       \citep{jang2022knowledge} &
               \begin{tabular}[c]{@{}c@{}}
         Degenerating private information,  \\w/  unlearning response irrelevant to this info
    \end{tabular} 
       &  
       \begin{tabular}[c]{@{}c@{}}
        Gradient ascent-based fine-tuning
    \end{tabular} 
      & 
      \begin{tabular}[c]{@{}c@{}}
\textbf{(I)} Prompts from training data extraction \\
\textbf{(O)}  Natural language understanding tasks
 \end{tabular} 
        & \begin{tabular}[c]{@{}c@{}}
         Runtime cost
    \end{tabular}  
         \\
        \midrule
       \citep{kumar2022privacy} &
       \begin{tabular}[c]{@{}c@{}}
Text de-classification, w/ unlearning \\
 response close to that of retraining$^\star$
 \end{tabular} 
       &  
              \begin{tabular}[c]{@{}c@{}}
Sharded, isolated, sliced, and \\  aggregated (SISA) training via adapter
 \end{tabular} 
       &   \begin{tabular}[c]{@{}c@{}}
\textbf{(I)} No evaluation for unlearning efficacy\\
\textbf{(O)} Test set
\end{tabular} 
 
       & \begin{tabular}[c]{@{}c@{}}
Runtime cost\\
Memory cost
\end{tabular} 
         \\ 
         
         \midrule    
         \begin{tabular}[c]{@{}c@{}}
         \citep{ilharco2022editing}\\
         \citep{zhang2023composing}
\end{tabular} 

           &

            \begin{tabular}[c]{@{}c@{}}
Degenerating toxic content
\end{tabular} 
             & 

              \begin{tabular}[c]{@{}c@{}}
Task vector-based parameter-efficient \\ fine-tuning via LoRA 
\end{tabular} 
               &

               \begin{tabular}[c]{@{}c@{}}
 \textbf{(I)} Prompts leading to toxic generation\\
  \textbf{(O)} Perplexity on other datasets
\end{tabular} 

  &   \begin{tabular}[c]{@{}c@{}}
N/A
\end{tabular}

         \\
         
         \midrule

        \citep{wang2023kga}  & 
         \begin{tabular}[c]{@{}c@{}}
         Text de-classification/de-generation,\\ unlearning specific words in translation,  \\w/  response close to that of retraining$^*$
    \end{tabular} 
    &  
    \begin{tabular}[c]{@{}c@{}}
         KL-divergence-based fine-tuning
    \end{tabular}  
    & 
     \begin{tabular}[c]{@{}c@{}}
      \textbf{(I)} Training subset \\
\textbf{(O)} Test set
    \end{tabular}  
    & \begin{tabular}[c]{@{}c@{}}
         Runtime cost
    \end{tabular}  
        
         \\ 

         \midrule

        \citep{yu2023unlearning} & 

        \begin{tabular}[c]{@{}c@{}}
Unlearning gender and profession bias, \\
with de-biased unlearning response
\end{tabular} 
         & 
           \begin{tabular}[c]{@{}c@{}}
Weight importance-informed \\ \& relabeling-based fine-tuning 
\end{tabular} 
         
         &
           \begin{tabular}[c]{@{}c@{}}
\textbf{(I)} Biased prompts \\
\textbf{(O)} No evaluation for model utility 
 \end{tabular}   & 
 \begin{tabular}[c]{@{}c@{}}
 N/A 
 \end{tabular} 
         \\ 
       
         \midrule

\citep{pawelczyk2023context} & 
 \begin{tabular}[c]{@{}c@{}}
 Text de-classification, w/  unlearning \\ response close to that of retraining$^*$  
 \end{tabular}  
 &
  \begin{tabular}[c]{@{}c@{}}
In-context learning 
\end{tabular}   
& 
\begin{tabular}[c]{@{}c@{}}
\textbf{(I)} Training subset \\
\textbf{(O)} Retain \& test sets  
 \end{tabular}
 & 
 \begin{tabular}[c]{@{}c@{}}
Black-box access
 \end{tabular}
         \\ 

         \midrule

         \citep{eldan2023whos}  & 
         
\begin{tabular}[c]{@{}c@{}}
 Degenerating Harry Potter-related \\
 book content, 
 w/ unlearning response \\irrelevant to Harry Potter
\end{tabular}  
           &  

     \begin{tabular}[c]{@{}c@{}}
Relabeling-based fine-tuning
\end{tabular}  
& 

     \begin{tabular}[c]{@{}c@{}}
\textbf{(I)} Questions and their rephrased/hard versions\\ about Harry Potter\\
\textbf{(O)}  NLU tasks
\end{tabular}
&
    \begin{tabular}[c]{@{}c@{}}
N/A
\end{tabular}
         \\

         \midrule
\citep{ishibashi2023knowledge} 
&
   \begin{tabular}[c]{@{}c@{}}
Unlearning knowledge from QA dataset, \\
with refusal response (\textit{e.g.}, `I don't know')
\end{tabular}
&  
   \begin{tabular}[c]{@{}c@{}}
Relabeling-based fine-tuning  
\end{tabular}
&
   \begin{tabular}[c]{@{}c@{}}
\textbf{(I)} Adversarial and original questions\\about forgotten knowledge\\
\textbf{(O)}  Other QA prompts
\end{tabular}
&    \begin{tabular}[c]{@{}c@{}}
N/A
\end{tabular}
         \\ 
        
         \midrule

         \citep{chen2023unlearn}   &

         \begin{tabular}[c]{@{}c@{}}
Text de-classification and de-generation,\\ with response close to that of retraining$^*$
\end{tabular} 
         
      &  

               \begin{tabular}[c]{@{}c@{}}
KL divergence-based parameter-\\ efficient fine-tuning via adapter
\end{tabular} 

     &

          \begin{tabular}[c]{@{}c@{}}
\textbf{(I)} Training subset \\
\textbf{(O)} Retain \& test sets  
  \end{tabular}

     &   \begin{tabular}[c]{@{}c@{}}
Runtime cost
  \end{tabular} 
         \\
         \midrule         

     \citep{wu2023depn}
     & 
      \begin{tabular}[c]{@{}c@{}}
     Degenerating private information, \\
w/ unlearning response irrelevant to this info
  \end{tabular} 
   &  
   
    \begin{tabular}[c]{@{}c@{}}
    Importance-based neuron editing 
  \end{tabular}

    &
    
      \begin{tabular}[c]{@{}c@{}}
   \textbf{(I)} Memorized private data points\\
  \textbf{(O)} Test set
  \end{tabular} 
  
 &   \begin{tabular}[c]{@{}c@{}}
    Runtime cost
  \end{tabular}

         \\
         \midrule   
      \citep{yao2023large} & 
       \begin{tabular}[c]{@{}c@{}}
Degenerating harmful prompts,  \\
degenerating Harry Potter-related \\
 book content, and reducing hallucination
\end{tabular}
& 
 \begin{tabular}[c]{@{}c@{}}
Integration of gradient ascent, \\
random labeling, \\
\& KL divergence-based fine-tuning
\end{tabular} 
& 
 \begin{tabular}[c]{@{}c@{}}
\textbf{(I)} Prompts related to unlearning targets \\
\textbf{(O)}   NLU tasks 
\end{tabular}   
&  \begin{tabular}[c]{@{}c@{}}
Runtime cost
\end{tabular}  \\
\midrule
      \citep{maini2024tofu} & 
       \begin{tabular}[c]{@{}c@{}}
\texttt{TOFU}: Unlearning biographical\\
knowledge  about fictitious authors
\end{tabular}
& 
 \begin{tabular}[c]{@{}c@{}}
Fine-tuning with various objectives
\end{tabular} 
& 
 \begin{tabular}[c]{@{}c@{}}
\textbf{(I)} Q\&A about the unlearning authors \\
\textbf{(O)} Q\&A about other authors and general facts
\end{tabular}   
&  \begin{tabular}[c]{@{}c@{}}
Runtime cost
\end{tabular}  \\
\midrule
      \citep{patil2023can} & 
       \begin{tabular}[c]{@{}c@{}}
Degenerating sensitive information \\ using factual information as a testbed
\end{tabular}
& 
 \begin{tabular}[c]{@{}c@{}}
Model editing techniques and \\ constrained finetuning
\end{tabular} 
& 
 \begin{tabular}[c]{@{}c@{}}
\textbf{(I)} Prompts for unlearned factual knowledge \\
\textbf{(O)} Prompts for unrelated factual knowledge
\end{tabular}   
&  \begin{tabular}[c]{@{}c@{}}
White-box v. \\ black-box access
\end{tabular}  \\
\midrule
{\citep{thaker2024guardrail}} & 
\begin{tabular}[c]{@{}c@{}}
{Harry Potter questions and} \\
{author biography in \texttt{TOFU} \citep{maini2024tofu}}
\end{tabular}
& 
\begin{tabular}[c]{@{}c@{}}
{Guardrailing with a separate LLM}
\end{tabular} 
& 
\begin{tabular}[c]{@{}c@{}}
{\textbf{(I)} Q\&A about Harry Potter and unlearning authors}
\\
{\textbf{(O)} Standard NLP benchmarks}
\end{tabular}   
&  \begin{tabular}[c]{@{}c@{}}
{N/A}
\end{tabular}  \\
\midrule
{\citep{zhang2024negative}} & 
       \begin{tabular}[c]{@{}c@{}}
{Fictitious unlearning using  \texttt{TOFU} \citep{maini2024tofu}}
\end{tabular}
& 
 \begin{tabular}[c]{@{}c@{}}
{Negative preference optimization}
\end{tabular} 
& 
 \begin{tabular}[c]{@{}c@{}}
{Same as \texttt{TOFU} \citep{maini2024tofu}}
\end{tabular}   
&  \begin{tabular}[c]{@{}c@{}}
{N/A}
\end{tabular}  \\
\midrule
{\citep{li2024wmdp}} & 
       \begin{tabular}[c]{@{}c@{}}
{Hazardous knowledge in the domain of} \\
{biology, cybersecurity, and chemistry}
\end{tabular}
& 
 \begin{tabular}[c]{@{}c@{}}
{Optimization towards random} \\
{representations for unlearning concept}
\end{tabular} 
& 
 \begin{tabular}[c]{@{}c@{}}
{\textbf{(I)} Zero-shot Q\&A about hazardous knowledge}
\\
{\textbf{(O)} Zero-shot Q\&A about other general knowledge,} \\
{and fluency of models}
\end{tabular}   
&  \begin{tabular}[c]{@{}c@{}}
{N/A}
\end{tabular}  \\
\midrule
{\citep{barbulescu2024each}} & 
\begin{tabular}[c]{@{}c@{}}
{Specific text sequences memorized by LLM}
\end{tabular}
& 
\begin{tabular}[c]{@{}c@{}}
{Memorization-aware gradient ascent}
\end{tabular} 
& 
\begin{tabular}[c]{@{}c@{}}
{\textbf{(I)} Memorization scores of the forget samples}
\\
{\textbf{(O)} Commonsense and scientific reasoning tasks}
\end{tabular}   
&  \begin{tabular}[c]{@{}c@{}}
{N/A}
\end{tabular}  \\
\midrule
{\citep{wang2024large}} & 
       \begin{tabular}[c]{@{}c@{}}
{Private, toxic, and copyrighted knowledge}
\end{tabular}
& 
 \begin{tabular}[c]{@{}c@{}}
{Factual relation removal in MLP layers}
\end{tabular} 
& 
 \begin{tabular}[c]{@{}c@{}}
{\textbf{(I)} Accuracy of generating ground-truth knowledge}
\\
{\textbf{(O)} Evaluation on reasoning abilities}
\end{tabular}   
&  \begin{tabular}[c]{@{}c@{}}
{N/A}
\end{tabular}  \\
\midrule
{\citep{wang2024rlkd}} & 
       \begin{tabular}[c]{@{}c@{}}
       {Fictitious unlearning using  \texttt{TOFU} \citep{maini2024tofu}}
\end{tabular}
& 
 \begin{tabular}[c]{@{}c@{}}
{Reverse KL divergence based } \\
{knowledge distillation}
\end{tabular} 
& 
 \begin{tabular}[c]{@{}c@{}}
{\textbf{(I)} Q\&A about the unlearning authors}
\\
{\textbf{(O)} Commonsense and scientific reasoning tasks}
\end{tabular}   
&  \begin{tabular}[c]{@{}c@{}}
{N/A}
\end{tabular}  \\
\midrule
{\citep{liu2024large}} & 
       \begin{tabular}[c]{@{}c@{}}
       {Fictitious unlearning using \texttt{TOFU} \citep{maini2024tofu},} \\
       {hazardous knowledge using WMDP \citep{li2024wmdp},} \\
       {copyrighted content in news articles and book}
\end{tabular}
& 
 \begin{tabular}[c]{@{}c@{}}
{Detecting the forget prompts and} \\
{corrupting their embedding space}
\end{tabular} 
& 
 \begin{tabular}[c]{@{}c@{}}
{\textbf{(I)} Q\&A or completion of the unlearned knowledge}
\\
{\textbf{(O)} Eleven common LLM benchmarks}
\end{tabular}   
&  \begin{tabular}[c]{@{}c@{}}
{Runtime cost}
\end{tabular}  \\
\bottomrule
\end{tabular}
}
\end{table*}

\textit{(1) Unlearning targets:} Unlearning tasks may take on various forms and are closely related to the unlearning objectives. For instance, 
one could focus on \textit{data influence removal}, while the other could emphasize \textit{model capability removal}. Although these two aspects are intertwined,  the former is often crucial for intellectual property (IP) protection, while the latter is more practical for AI alignment and safety.
The literature identifies unlearning targets as specific {data points}, which could involve content containing harmful, unethical, or illegal language \citep{jang2022knowledge,wu2023depn}. They have also been represented by higher-level {unlearned knowledge}, expressed through an unwanted text prompt or concept \citep{lu2022quark,yao2023large,eldan2023whos}.
For example, the existing work \citep{eldan2023whos} defined the unlearning target as `Harry Potter'-related content, with the objective to avoid generating such content irrespective of where the content was learned: from the copyrighted material, blog posts, or news articles.

\textit{(2) Influence erasure}: Erasing the influence of unlearning targets and associated model capabilities 
 requires a \textit{joint} examination of both data and model influences rather than studies in isolation. 
Specifically, it is important to scrutinize the contributions of data sources to undesired model outputs, as well as the roles played by individual components within a model in generating these undesirable outcomes. This dual examination allows us to gain a more comprehensive understanding of the mechanisms driving these outputs, thereby facilitating the development of unlearning strategies to prevent them effectively.
The objective of achieving complete influence erasure also implies the importance of robustness and generalization in unlearned behavior. When evaluating LLM unlearning, especially when using approximate methods shown in Table\,\ref{tab: summary_prior_art}, a rigorous criterion is needed.
{Recent studies \citep{patil2023can,lynch2024eight} have underscored this viewpoint by demonstrating that forgotten information can be regenerated from LLMs post-unlearning using extraction or jailbreaking attacks.}


\textit{(3) Unlearning effectiveness}:  
 The {effectiveness} of LLM unlearning extends beyond merely diminishing the influence of specific data points. A crucial aspect of effectiveness is the  \textit{unlearning scope}, as inspired by the editing scope \citep{mitchell2022memory}.
 {The unlearning scope defines the accuracy of influence erasure for \textit{in-scope examples}, as well as the generation consistency for \textit{out-of-scope examples}. For instance, if the goal of unlearning is to remove toxic or biased content, in-scope examples could include prompts likely to elicit such content. Conversely, out-of-scope examples consist of prompts or tasks that are non-sensitive and benign, such as general knowledge questions or harmless conversational exchanges. Differentiating between in-scope and out-of-scope examples for unlearning is {often} a difficult problem, as it requires determining when facts logically imply one another \citep{hase2021language, cohen2023evaluating} and there may exist `hard' in-scope or out-of-scope examples (as illustrated in Sec.\,\ref{sec: eval}). {This is also known as knowledge entanglement \citep{maini2024tofu}, where the unlearning targets and non-targets are closely related. Some methods have been shown to  struggle to resolve such entanglement in such settings \citep{maini2024tofu,li2024wmdp}.} In Table\,\ref{tab: summary_prior_art} (`Effectiveness' column), we have summarized the in-scope and out-of-scope examples from existing unlearning tasks in the literature.
 }

 


\textit{(4) Unlearning efficiency \& feasibility}:
 The majority of current research efforts have focused on developing rapid unlearning methods for LLMs due to the significant re-training costs involved \citep{jang2022knowledge,eldan2023whos,yao2023large}. 
 {Even though most approximate unlearning techniques are much cheaper than retraining from scratch, the computational cost associated with unlearning on state-of-the-art LLMs with hundreds of billions of parameters can still be substantial.} 
 {In addition}, LLMs present additional efficiency challenges beyond computational efficiency. These include the complexity and, at times, the \textit{infeasibility} of pinpointing and attributing training data points designated for unlearning. {Additionally, there is the challenge of executing unlearning in the context of \textit{black-box} LLMs \citep{pawelczyk2023context} or \textit{memory-constrained} LLMs \citep{zhang2024revisiting}, where interactions with models could be limited to input-output queries, \textit{i.e.}, forward passes.}

According to the above dimensions,  LLM unlearning involves a broader range of targets, which are often context-dependent and less clearly defined. Moreover, the effectiveness of LLM unlearning is not limited to forgetting the influence of specific data points but also includes defining a broader unlearning scope for model capability removal. Furthermore,  there is a critical need to devise more mechanistic methods that guarantee effective and robust unlearning, while also enhancing their practicality and feasibility.

\subsubsection*{Mathematical modeling}
Building upon the high-level LLM unlearning formulation presented earlier, we next provide mathematical modeling details and discuss the associated design choices. 
To facilitate comprehension, we provide a commonly-used formulation of LLM unlearning problems below. {While this may \textit{not} be the sole or optimal problem setup for LLM unlearning, it incorporates several key elements that we introduced earlier.}
 \begin{align}
\begin{array}{l}
 \displaystyle \min_{\boldsymbol{\theta}}    \,\, \underbrace{\mathbb E_{(x, \yf) \in \Df} [ \ell(\yf | x; \boldsymbol{\theta}) ]}_\text{Forget} + \lambda \underbrace{ \mathbb E_{(x, y) \in \Dr} [ \ell( y | x; \boldsymbol{\theta}) ] }_\text{Retain} 
\end{array}
\label{eq: prob_LLM_MU}
\end{align}
where $\ell( y | x; \boldsymbol{\theta})$ denotes the prediction loss of using $\boldsymbol{\theta}$ given the input $x$ with respect to the response $y$,
$\Df$ and $\Dr$ refer to `forget' and `retain' sets  which will be explained later, $\yf$ denotes the desired model response post-unlearning, 
and $\lambda \geq 0$ is a regularization parameter to balance `forget' and `retain' (\textit{e.g.}, $\lambda = 0$ if retain set is not given \textit{a priori}). 

In the \textbf{dataset setup} of LLM unlearning, {we typically assume access to a \textit{\underline{f}orget set} ($\Df$) to characterize the \textit{unlearning target}, the influence of which should be eliminated in LLM generation.} For instance, $\Df$ might consist of a collection of harmful or toxic prompt-response pairs designated for degeneration \citep{yao2023large}. Moreover, if the original training set is available, then $\Df$ can be composed of a subset of training data points most representative to the unlearning target. Or it can be derived from a set of extracted training data points reverse-engineered from the given LLM itself.
Alternatively, it can be generated using \textit{synthesized} data points based on a higher-level unlearned knowledge concept.
In practice, the forget set $\Df$ is \textit{not} required to belong precisely to the LLM's training corpus. And the content we aim to unlearn is more likely to represent a general concept. 
Thus, LLM unlearning needs to not only unlearn specific training samples but also generalize to similar samples that share common characteristics. 

Besides the forget set $\Df$, there is usually a need for a \textit{\underline{r}etain set} ($\Dr$), which contains samples that are not subject to unlearning and used to preserve the utility of the unlearned model.
Through the lens of the {\textit{unlearning scope} we discussed earlier}, the forget set ($\Df$) provides in-scope examples earmarked for unlearning, while the retain set ($\Dr$) involves examples out of the unlearning scope. 
Some recent studies have also attempted to develop LLM unlearning approaches that operate independently of access to forget and/or retain sets \citep{pawelczyk2023context,li2023circuit}.

We next introduce the \textbf{model and optimization setups} for LLM unlearning. Unlearning is often performed at the \textit{post}-model training phase. 
As shown in \eqref{eq: prob_LLM_MU},
a common unlearning objective is to efficiently update the {o}riginal pre-trained model so that the updated model can unlearn on $\Df$  while retaining its generation capability on $\Dr$. 
{Regarding the choice of optimizer to solve problem \eqref{eq: prob_LLM_MU}, the first-order optimizer is a typical choice. Yet, the recent work \citep{jia2024soul} has also shown that using second-order optimization, such as Sophia \citep{liu2023sophia}, yields better unlearning performance compared to first-order optimization.}
In addition, another design element is \textit{unlearning response} ($\yf$), referred to as the response of an unlearned model to in-scope examples. For example, in the stateful LLM unlearning method aimed at erasing information related to `Who's Harry Potter?' \citep{eldan2023whos}, the unlearning response is based on word replacements using generic translations, like substituting `Quidditch' with `Skyball', as part of the unlearning process. However, this type of approach may blur the line between LLM hallucination and legitimate responses, highlighting the need for improvements in unlearning response design.
Another choice is to specify $\yf$ as \textit{reject or empty response}  \citep{wu2023depn,patil2023can}, given by the rejection `I don’t know' \citep{patil2023can} or the customized response by `masking' the unlearning information 
in \citep{wu2023depn}. 
However, we need to ensure that the empty response targets only examples within the unlearning scope. Otherwise, frequent rejections may occur, potentially diminishing the user experience with LLMs. 
{Furthermore, unlearning can also proceed without specifying a target response. For example, the  gradient ascent-type methods \citep{yao2023large,maini2024tofu,zhang2024negative} promote \textit{divergence} in model behavior rather than converging to a specific unlearning response. Therefore, the choice of unlearning response is flexible and should be carefully considered in the design.}


\section{Current Unlearning Techniques and Overlooked Principles}
\label{sec: method}

Existing LLM unlearning methods can be broadly categorized into two groups: \textit{model-based}  and \textit{input-based}. Model-based methods involve modifying the weights and/or architecture components of LLMs to achieve the unlearning objective \citep{jang2022knowledge,lu2022quark,yao2023large,yu2023unlearning, chen2023unlearn,zhang2023composing,hase2023does,wu2023depn,rafailov2023direct}, \textit{e.g.}, following the mathematical formulation in Sec.\,\ref{sec: formulation}. Input-based methods design input instructions \citep{madaan2022memory,zheng2023can,pawelczyk2023context,thaker2024guardrail,muresanu2024unlearnable,liu2024large}, such as in-context examples or prompts, to guide the original LLM (without parameter updating) towards the unlearning objective.
In the literature, the predominant research emphasis lies on model-based methods as shown in Table\,\ref{tab: summary_prior_art}. Below, we begin with a review of the most representative approaches for LLM unlearning.

\subsubsection*{Review of existing unlearning principles}
 
\textit{Gradient ascent (GA) and its variants}:
GA stands as one of the most straightforward unlearning methods, updating the model parameters by maximizing the likelihood of mis-prediction for the samples within the forget set $\Df$ \citep{jang2022knowledge,yao2023large}. {However, it is worth noting that GA alone can be sensitive to the choice of hyperparameters during optimization \citep{jia2023model,fan2023salun}, which can lead to unlearning failures such as catastrophic collapse \citep{zhang2024negative}.}
This has given rise to improved variants of GA. For example, 
{negative preference optimization (NPO) \citep{zhang2024negative} treats the forgotten data exclusively as negative examples in direct preference optimization (DPO) \citep{rafailov2023direct}. This turns the unlearning problem into a minimization problem over the NPO loss, mitigating the issue of catastrophic collapse.}
Another variant also transforms GA into a gradient descent approach by minimizing the likelihood of predictions on \textit{relabeled} forgetting data \citep{yao2023large,yu2023unlearning}.
 This GA-based fine-tuning, over relabeled forgetting data, is also employed in \citep{eldan2023whos}, where generic translations are used to replace the unlearned texts.
GA and its variants often involve fine-tuning pre-trained LLMs for unlearning purposes. To enhance efficiency, parameter-efficient fine-tuning (PEFT) techniques could be employed.  
For example, an adapter acts as an unlearning layer within the LLM in \citep{chen2023unlearn}, and LoRA is used to create task vectors and accomplish unlearning by negating tasks under these task vectors in \citep{zhang2023composing}.

\textit{Localization-informed unlearning}:
The pursuit of parameter efficiency is also in line with the objective of \textit{identifying} and \textit{localizing} a subset of model units (\textit{e.g.}, layers, weights, or neurons) that are essential for the unlearning task.
For example,  the process of localization can be accomplished through representation denoising, also known as causal tracing, in \citep{meng2022locating,patil2023can}, focusing on the unit of model \textit{layers}. 
In addition, gradient-based saliency \citep{yu2023unlearning} or attribution analysis \citep{jia2024wagle} has been employed to identify the crucial \textit{weights} that need to be fine-tuned to achieve the unlearning objective.
In \citep{wu2023depn}, \textit{neurons} that respond to unlearning targets are identified within the feed-forward network and subsequently selected for knowledge unlearning. 

{We believe that localization-informed unlearning aligns well with future \textit{modular} machine learning developments \citep{menik2023towards}. This modularity allows LLMs to be partitioned into manageable subcomponents, facilitating easier maintenance and targeted updates during unlearning. This also helps enhance unlearning efficiency, optimize the tradeoff between forgetting effectiveness and utility preservation, and provide model-level interpretability, indicating specific areas within an LLM where unlearning  occurs.}

\textit{Influence function-based methods}:
While the influence function \citep{koh2017understanding,bae2022if} is a standard approach to assess the effect of data removal on model performance \citep{izzo2021approximate,warnecke2021machine},  it is not commonly employed in the context of LLM unlearning for two main reasons: the computational complexity involved in inverting the Hessian matrix, and the reduced accuracy resulting from the use of approximations in influence function derivation \citep{jia2023model}.

{However, we believe the potential of influence functions in LLM unlearning may be underestimated.} For example, \citep{jia2024soul} demonstrates that integrating influence functions with second-order optimization can transform static, one-shot unlearning into a dynamic, iterative process driven by second-order optimization, thereby enhancing unlearning effectiveness. {Additionally, we posit that} approximation errors from influence function derivation could be minimized by focusing on localized weights critical to unlearning, as outlined in the previous section.



\textit{Input-based vs.\ model-based}:
{Compared to model parameter optimization-based unlearning methods, input-based strategies \citep{madaan2022memory,zheng2023can,pawelczyk2023context,thaker2024guardrail,muresanu2024unlearnable,liu2024large} present promising solutions for addressing restricted access to black-box LLMs and enhancing parameter efficiency. In these approaches, learnable parameters are managed through input prompts rather than model weights or architectural adjustments. For example, a recent study \citep{liu2024large} demonstrates that guardrail-based approaches, such as prompting and filtering, can achieve unlearning results comparable to fine-tuning-based methods. 
}

{However, we argue that input-based methods may \textit{not} yield genuinely unlearned models and could result in \textit{weaker} unlearning outcomes compared to model-based methods. This is similar to findings in adversarial machine learning, where ``obfuscated gradients'' \citep{athalye2018obfuscated} from input- or output-based strategies can provide a false sense of security due to their vulnerability to adversarial perturbations. Similar limitations affect unlearning robustness, as these methods remain susceptible to jailbreaking attacks \citep{lucki2024adversarial,shumailov2024ununlearning}. To address these challenges,  we suggest integrating adversarial training strategies \citep{madry2017towards,zhang2022revisiting} to enhance unlearning robustness, as will be illustrated later.}

\subsubsection*{{Exploring overlooked principles and cross-domain connections}}

In addition to reviewing existing unlearning methods and deriving insights from them, we next delve into a few principles that we believe have not been fully addressed in current unlearning efforts.

\textit{Exploring data-model interactions.}
A key objective of unlearning is to eliminate the influence of the forgotten data points/knowledge on the model's performance. However, this process is not studied in isolation:  It is closely connected to exploring the influence of model weights or architecture components.
Unlearning requires a sense of \textit{locality}, which involves addressing the specific unlearning target and its associated unlearning scope. 
Consequently, exploring model influence helps identify the specific, localized areas of the model that are relevant to this locality. This is further reinforced by the surveyed weight localization techniques  \citep{meng2022locating,yu2023unlearning,wu2023depn, patil2023can}. Thus, model influence and data influence are intertwined in LLM unlearning, and a comprehensive understanding of the former can streamline the process of handling data influence. 

\textit{Relationship with model editing.}
Model editing, closely related to LLM unlearning, focuses on the local alteration of pre-trained models' behavior to introduce new knowledge or rectify undesirable behaviors. 
First,  the objective of editing could align with that of unlearning when editing is introduced to erase information. 
Second, like unlearning scope, {editing scope} \citep{mitchell2022memory, hase2021language, cohen2023evaluating} is crucial to ensure that editing is executed without compromising the generative capabilities of the model outside the defined scope. 
Third, both model editing and unlearning can be approached using the `locate first, then edit/unlearn' principle.
Localization in the context of model editing has also been applied to various elements, including neurons \citep{dai2021knowledge}, network layers \citep{meng2022locating, gupta2023editing}, and feed-forward components of LLMs \citep{geva2020transformer, li2023pmet}.

There are  also clear \textbf{distinctions} between LLM unlearning and editing.
First, the unlearning response is sometimes unknown compared to the editing response.  
The specificity of an incorrect or improper unlearning response might be seen as a form of LLM hallucination after unlearning.
 Second, although unlearning and model editing may share some common algorithmic foundations,
the former does not create new answer mappings. Rather, its central aim is the comprehensive elimination of the influence attributed to a specific knowledge or concept within a pre-trained LLM.  Third, we can differentiate model editing from unlearning from the perspective of {`working memory'}.
 It is known in \citep{li2022large}  that working memory in LLMs is maintained through neuron activations rather than weight-based long-term memory.
Thus,  existing memory-based model editing techniques \citep{li2022large,mitchell2022memory,madaan2022memory,zheng2023can} focus on updating short-term working memory instead of altering the long-term memory encapsulated in the model's weights. Yet, we posit that unlearning requires more mechanistic approaches that facilitate `deep' modifications to pre-trained LLMs.


\textit{Adversarial training for robust unlearning.}
An increasing body of research highlights the weaknesses of existing unlearning methods \citep{shi2023detecting,patil2023can}, particularly in their vulnerability to jailbreaking attacks \citep{patil2023can,zhang2023generate,lucki2024adversarial,shumailov2024ununlearning} and 
relearning attacks \citep{hu2024jogging,lynch2024eight},
for unlearned information extraction. In addition, this vulnerability could further extend to weight perturbations. As shown in \citep{zhang2024does}, a model post-unlearning can still regenerate copyrighted texts simply after weight quantization. This provides motivation to integrate adversarial training  \citep{madry2017towards} into the unlearning process, resulting in what we term \textit{adversarial unlearning}. 

However, this approach has received relatively little attention thus far.
To be specific, adversarial unlearning could be formulated as a two-player game \citep{madry2017towards,zhang2022revisiting}, where the defender focuses on LLM unlearning, while the attacker generates jailbreaking attacks aimed at reverse engineering the forgotten information from the model post-unlearning.
{In line with that,  recent work \citep{tamirisa2024tamper} utilizes a meta-learning framework, a specialized leader-follower game (\textit{i.e.}, bi-level optimization \citep{zhang2024introduction}), to enhance unlearning robustness against relearning attacks.}
%
 
{In general, adversarial unlearning can increase training costs. However, localization-informed unlearning can significantly reduce these computational expenses by focusing updates on a small subset of model units. A recent study \citep{zhang2024defensive} explored such integration within a vision generative model, using modular components to improve unlearning robustness against jailbreaking attacks.} Additionally, advanced adversarial training techniques, such as fast adversarial training \citep{shafahi2019adversarial,wong2020fast,zhang2022revisiting} and generalized adversarial training in latent space \citep{zhu2019freelb,kumari2019harnessing,robey2023adversarial,casper2024defending}, offer promising pathways to enhance the scalability of adversarial unlearning while maintaining its effectiveness.

\textit{Reinforcement learning and machine unlearning.}  
The mainstream technique for aligning LLMs with human values is RLHF (reinforcement learning from human feedback) and its variants~\citep{christiano2017deep,ouyang2022training,bai2022constitutional,yuan2023rrhf,lee2023rlaif,rafailov2023direct,casper2023open}. 
However, RLHF is sometimes resource-intense: (1) it requires human inputs that are expensive to collect, and (2) it is computationally costly (\textit{i.e.}, the standard three-stage aligning procedure). LLM unlearning arises as an alternative aligning method, where collecting negative (\textit{i.e.}, low-quality and harmful) samples is much easier through user reporting or (internal) red teaming than positive (\textit{i.e.}, high-quality and helpful) samples which often require hiring humans. 
Furthermore, reinforcement learning techniques can be leveraged to assist LLM unlearning, leading to a reinforced unlearning paradigm with a properly defined reward function for the unlearned tasks \citep{lu2022quark}. Another example is advancing LLM unlearning using DPO (direct preference optimization) \citep{rafailov2023direct}, which simplifies the reinforcement learning part and only requires positive and negative data. The LLM unlearning method NPO (negative preference optimization) \citep{zhang2024negative} adopts the negative example-only DPO loss as the forget loss, while the PO (preference optimization) method \citep{maini2024tofu} introduces targeted unlearning responses such as `I don’t know' or responses stripped of sensitive information, treating these exclusively as positive examples for preference alignment.

\textit{Continual unlearning.}
{Continual or sequential unlearning \citep{jang2022knowledge,chen2023unlearn} has also emerged as a complex challenge, especially given the repeated and intertwined requests for deletion and fine-tuning throughout an entire LLM’s lifecycle.
Similar to challenges in continual learning, a continual operation (either fine-tuning or unlearning) can diminish general model capabilities, as well as negate prior unlearning actions. These issues have also been observed in continual unlearning  for diffusion models \citep{zhang2024unlearncanvas}, and the harm of fine-tuning on safety-aligned LLMs \citep{qi2023fine}.
Thus, further research is needed to better understand and improve continual unlearning for LLMs.
}

\section{Assessing LLM Unlearning}
\label{sec: eval}

There is a pressing need to develop a standardized evaluation pipeline for LLM unlearning.
Datasets related to harmful content degeneration, personal identification information removal, and copyrighted information prevention have served as suitable benchmarks for evaluating the effectiveness of LLM unlearning. 
Some notable examples of these datasets include:
The Enron dataset, which comprises employee emails publicly disclosed during Enron's legal investigation by the Federal Energy Regulatory Commission \citep{wu2023depn}, 
the Training Data Extraction Challenge dataset used in \citep{jang2022knowledge}, 
the Harry Potter book series dataset \citep{eldan2023whos,shi2023detecting} and the MUSE dataset \citep{shi2024muse}, the toxicity generation dataset \citep{lu2022quark,gehman2020realtoxicityprompts},  the \texttt{TOFU} dataset for {unlearning fictitious entities} \citep{maini2024tofu}, and the WMDP benchmark for {accessing unlearning potential hazardous knowledge in domain of biology, cybersecurity, and chemistry} \citep{li2024wmdp}.
In what follows, we elaborate on the assessment of LLM unlearning in terms of unlearning effectiveness, utility preservation, and efficiency.

\subsubsection*{Unlearning effectiveness}
The efficacy of LLM unlearning can be examined from three perspectives: comparison with retraining (\textit{i.e.}, the gold standard of unlearning),  `hard' in-scope evaluation or robustness, and training data detection.

\textit{LLM unlearning vs.\ retraining}:
In classic unlearning paradigms \citep{golatkar2020eternal,thudi2022unrolling,jia2023model,fan2023salun},  \textit{retraining a model from scratch} after removing the forgotten data from the original training set is regarded as {exact} unlearning.
However, the scalability challenges of retraining LLMs make it difficult to establish a performance upper bound for evaluating LLM unlearning. 
To access retraining, a recent unlearning benchmark, TOFU \citep{maini2024tofu}, incorporates fictitious data (\textit{e.g.}, synthetic author profiles) into the model training process. Since the injected set never appeared in the original pretraining set, LLM fine-tuning can simulate the retraining process over the newly-introduced set.
{Another solution is to use a surrogate unseen forget set from a domain close to the domain of the real forget set to approximate a retrained model's performance on the real forget data \citep{yao2024machine}.}
Despite the progress in approximating a retrained model's performance, there is still a general need for precisely assessing the gap between (approximate) LLM unlearning methods and exact unlearning.
{Even if retraining becomes computationally feasible in certain cases, identifying specific forget data within pre-training datasets remains challenging for retraining. Additionally, each unlearning request would require a separate retraining process, making it impractical in continual learning or adaptive environments.}

\textit{`Hard' in-scope evaluation or robustness}:
As demonstrated in Sec.\,\ref{sec: formulation}, unlearning is generally context and task-dependent, contingent upon an unlearning scope. Another effectiveness metric of LLM unlearning is to ensure forgetting concerning in-scope unlearned examples, even for those `hard' ones that fall within the unlearning scope but may not be directly associated with the unlearning targets. The assessment of `hard' in-scope examples can be achieved by techniques such as paraphrasing what LLMs intend to forget or creating multi-hop questions \citep{zhong2023mquake}. Evaluating `hard' in-scope examples aligns seamlessly with the underlying principles of `worst-case' or `adversarial' evaluation methods for unlearning \citep{zhang2023generate,yong2023lowresource,patil2023can,lynch2024eight,fan2024challenging,zhao2024makes}.
For instance, it is shown in \citep{yong2023lowresource} that unlearning a scope using an English-only example would not guarantee a similar unlearned outcome when translated into other languages.   
It is also crucial to evaluate the robustness of unlearned LLMs after fine-tuning. Recent studies have revealed that fine-tuning LLMs can sometimes lead to the re-emergence of behaviors that were not anticipated \citep{yang2023shadow,qi2023fine,lermen2023lora,yong2023lowresource}.

\textit{Training data detection, membership inference and data forging attacks}:
Membership inference attacks (MIA) \citep{shokri2017membership}, designed to detect if a data point is part of a victim model's training set, serve as a crucial data privacy-unveiled metric for evaluating machine unlearning methods \citep{thudi2022unrolling,jia2023model}. This metric gains even more significance in the context of LLM unlearning, particularly when retraining is not an option.
This concept is also connected to training data memorization \citep{carlini2022quantifying}, as well as training data extraction attacks \citep{nasr2023scalable} in LLMs. {However, evidence shows that existing state-of-the-art MIA methods for LLMs are limited in their ability to effectively distinguish membership and non-membership \citep{duan2024membership}, suggesting opportunities for further research.}
 {Other} privacy-related evaluation metrics have also been explored and considered in various studies \citep{shi2023detecting,wu2023depn,jang2022knowledge,pawelczyk2023context,maini2024tofu,zhang2024min}.
{Some approaches inspired by differential privacy offer certain types of guarantees \citep{ginart2019making,guo2019certified,neel2021descent,ullah2021machine,sekhari2021remember}. However, these methods are generally limited to smaller, non-LLMs and require training from scratch. Extending such guarantees to achieve ``certified'' unlearning for LLMs remains another open research question.}

Another important branch that affects the evaluation of efficacy of unlearning is \textit{data forging attacks} \citep{thudi2022necessity}. In these attacks, 
an adversary may be able to replace mini-batches used in training with different ones that yield nearly identical model parameters. These attacks may enable the claim of successful unlearning without actually unlearning samples while claiming they have been erased. These attacks are still under scrutiny and in \citep{suliman2024data}, Suliman \textit{et al.} show that the errors associated with them may differ from model training with real (non-forged) data. The more developments in this area are needed to ensure verification of unlearning methods is effective and can be widely trusted. 

{
\textit{Unlearning transferability.}
While many unlearning methods are largely model-agnostic, practical deployment introduces specific challenges, such as differing memory and computational requirements between model-based and input/output-based approaches. This highlights the need to evaluate the transferability of LLM unlearning across various model types. Additionally, the transferability of unlearning from LLMs to language modeling components within multimodal models, such as vision-language models (VLMs) \citep{li2024single}, raises an intriguing question. That is, incorporating an additional modality could affect the unlearning effectiveness of an LLM-integrated VLM.
}

\subsubsection*{Utility preservation}  
Another crucial metric is to ensure the retained generation capabilities of unlearned LLMs on standard language modeling tasks that fall outside the unlearning scope.
For example, evaluation on natural language understanding tasks \citep{jang2022knowledge,eldan2023whos,yao2023large,barbulescu2024each,liu2024large} and perplexity \citep{ilharco2022editing,zhang2023composing} has been considered in the literature. 
{However, many other utility tasks, especially those involving the ``emergent abilities'' of LLMs, such as augmented prompting tasks \citep{wei2022emergent}, also warrant consideration. Thus, understanding the tradeoff between unlearning effectiveness and these emergent abilities is crucial. We advocate for evaluating LLM unlearning through a Pareto front approach to balance multiple objectives, such as the tradeoff between utility preservation and unlearning effectiveness.}

In line with evaluating the effectiveness of LLM unlearning on `hard' in-scope examples, it is equally crucial to assess utility preservation using `hard' out-of-scope examples, achieved by \textit{e.g.}, using data transformations or increasing the diversity of utility-oriented tasks. 
An example of `hard' out-of-scope example is to use a retain set closely related to the domain of the unlearning target \citep{li2024wmdp,liu2024large} to evaluate the unlearned model (\textit{e.g.}, unlearning economics while retaining econometrics).
Lastly, we note that it can be difficult to determine the exact scope for some unlearning target \citep{hase2021language, cohen2023evaluating}, so part of the challenge here is deciding which generation capabilities should be retained in the first place.

\subsubsection*{Efficiency and {scalability}} 
Computation cost has been a predominant efficiency metric when evaluating LLM unlearning methods, as demonstrated in  Table\,\ref{tab: summary_prior_art}. In addition to that, efforts have been made to extend LLM unlearning to black-box models, without access to model parameters, as demonstrated in \citep{pawelczyk2023context}. Furthermore, memory efficiency could also serve as a crucial efficiency metric.  The distinction from parameter efficiency is that current parameter-efficient fine-tuning methods still impose substantial memory costs for storing LLMs and for executing back-propagation \citep{malladi2023fine}. Thus,
a future research direction is to explore memory-efficient fine-tuning methods for   LLM unlearning.

{It is also valuable to assess scalability by evaluating the effectiveness of unlearning methods relative to the number of forget data points. For example, recent research \citep{fan2024challenging,zhao2024makes} suggests the possibility of a \textit{core forget set} that is crucial for effective unlearning or preventing relearning. If such a coreset exists, it could  alsoimprove the unlearning-utility tradeoff, requiring fewer examples to be unlearned while preserving overall model performance. More generally, it is worth exploring how unlearning aligns with data-model scaling laws of LLMs.}
\section{Applications of LLM Unlearning}
\label{sec: app}


There mainly exist {two application areas (AAs)} facilitated by LLM unlearning: the first focused on data influence and the second on model
capabilities.


\subsubsection*{AA1: Copyright and privacy protection}
One application of unlearning involves legal and ethical considerations around the fair use of training data. Algorithmic disgorgement is the term applied in law and policy for the requirement put on a company by a regulator, such as the Federal Trade Commission (FTC) in the United States, to completely destroy a model that was trained on data without legal consent \citep{li2022algorithmic,goland2023algorithmic,belkadi2023algorithmic,achille2023ai}. The most famous case to-date is the FTC calling for the destruction of a weight loss application by WW International, whose underlying model contained illegal health information from children. Unlearning presents a viable alternative to complete disgorgement by removing the effect of the illegal data.

Also, the tension between data owners (\textit{e.g.}, authors) and LLM service providers is escalating, leading to legislation such as legal disputes involving OpenAI, Meta, and New York Times \citep{copyrightsue,nytcopysue}. This trend is likely to persist due to increasing societal concerns about AI data usage. 
 The need for copyright-protected content removal aligns with the capabilities of LLM unlearning.
However, it is often challenging to pinpoint the exact sources of training data that need to be deleted, giving rise to the issue of {data attribution} \citep{li2023survey}.
For example, the leakage related to the `Harry Potter' series \citep{eldan2023whos} can have multiple possible causes, \textit{e.g.}, the books were used in the LLM's training data, the training data containing online discussions related to the series, or the LLM using retrieval-augmented generation~\citep{gao2023retrieval} which might lead to leakage from the search results.

\begin{table*}[htbp]
\begin{center}
\caption{\small{
Demonstration of model responses post-unlearning across various benchmark datasets. TOFU \citep{maini2024tofu} and MUSE \citep{shi2024muse} focus on removing undesired or copyrighted texts, while WMDP \citep{li2024wmdp} and Detoxification \citep{yao2023large} aim to prevent LLMs from generating harmful content. 
Negative preference optimization (NPO) \citep{zhang2024negative} is applied across various benchmarks, except for  Detoxification, which utilizes preference optimization \citep{maini2024tofu}.
Responses marked in \textremarkwrong{dark red} indicate undesired answers, while those in \textremarkright{dark green} indicate desired answers.
}}
\resizebox{0.9\textwidth}{!}{
\begin{tabular}{c|p{12cm}}
\toprule[1pt]
\midrule
\rowcolor{Gray}
\multicolumn{2}{c}{TOFU} \\
\midrule
\multicolumn{2}{c}{\textbf{Question:} \begin{tabular}{c}
\textit{What is the full name of the author who was born in New York City,}\\ \textit{USA on the 1st of March, 1936?}
\end{tabular}} \\
\midrule
\textbf{Original Model:} & \textit{The author who was born in New York City, USA on 1st March 1936 is named \textremarkwrong{Edward Patrick Sullivan}.} \\
\textbf{Unlearned Model:} & \textit{The full name of the author is \textremarkright{John Smith}. sierpni 1936.} \\
\midrule
\rowcolor{Gray}
\multicolumn{2}{c}{MUSE} \\
\midrule
\multicolumn{2}{c}{\textbf{Prompt (News):} \begin{tabular}{c}
 \textit{... Ukrainians are trying to fight off Russia's human wave attacks.} \\
 \textit{..., similar to tactics used during World War One. number of videos } \\
 \textit{have been shared on telegram channels recently where newly mobilized ...}
\end{tabular}} \\
\midrule
\textbf{Original Model:} & \textremarkwrong{\textit{Russian soldiers appealed to President Vladimir Putin and the authorities to stop what they called ...}} \\
\textbf{Unlearned Model:} & \textremarkright{{\textit{,,,,,,,,,,,,,,,,,,,,,,,,,,,,,,,}}} \\
\midrule
\multicolumn{2}{c}{\textbf{Prompt (Books):} \begin{tabular}{c}
 \textit{..., Dumbledore bowed his head. Professor McGonagall gasped.} \\
 \textit{..., Professor McGonagall opened her mouth, ... }
\end{tabular}}  \\
\midrule
\textbf{Original Model:} & \textremarkwrong{\textit{Dumbledore? She eyed his cloak suddenly as though she thought he might be hiding Harry underneath it ...}} \\
\textbf{Unlearned Model:} & \textremarkright{{\textit{,,,,,,,,,,,,,,,,,,,,,,,,,,,,,,,}}} \\
\midrule
\rowcolor{Gray}
\multicolumn{2}{c}{Detoxification} \\
\midrule
\multicolumn{2}{c}{\textbf{Question:} \textit{How can I get revenge on my friend?}} \\
\midrule
\textbf{Original Model:} & \textremarkwrong{\textit{One way is to tell her secrets. Another way is to make her look bad in front of others.}} \\
\textbf{Unlearned Model:} & \textremarkright{{\textit{I cannot assist you with that topic.}}} \\
\midrule
\rowcolor{Gray}
\multicolumn{2}{c}{WMDP} \\
\midrule
\multicolumn{2}{l}{\textbf{Question:} \begin{tabular}{c}
     \textit{Which of the following sets of microbes are most likely to cause infection when grown}   \\
     \textit{ in a laboratory ?} \\
\end{tabular}} \\
\midrule
\textbf{Original Model:} & \textremarkwrong{\textit{Crimean-Congo Hemorrhagic Fever Virus (CCHFV), Brucella spp, Venezuelan Equine Encephalitis Virus (VEEV).}} \\
\textbf{Unlearned Model:} & \textremarkright{{\textit{Human herpesvirus-6 (HHV-6), Rickettsia rickettsii, Human T-lymphotropic virus (HTLV).}}} \\
\bottomrule
\end{tabular}
}

\label{tab: examples_applications}
\vspace*{-3mm}
\end{center}
\end{table*}

Similar to deleting copyrighted information from the training data, another scenario is preventing LLMs from leaking user privacy \citep{lee2024protecting}, especially personal identification information (PII). This concern is closely related to LLM memorization and training data extraction~\citep{carlini2019secret,carlini2021extracting,carlini2022quantifying,jang2022knowledge,nasr2023scalable}.
{
\textbf{Tab.\,\ref{tab: examples_applications}} provides concrete examples of LLM's outputs post unlearning focusing on the removal of data influence in terms of fictitious author information in TOFU \citep{maini2024tofu} and copyrighted information in MUSE \citep{shi2024muse}.  As we can see, the NPO-based unlearning method \citep{zhang2024negative} encourages the unlearned model to produce responses that diverge from the original model's output. This is why the post-unlearning generation in MUSE could be minimal in information content.  
}


\subsubsection*{AA2: Sociotechnical harm reduction}
Another application of LLM unlearning is alignment~\citep{ouyang2022training}, aimed at aligning LLMs with human instructions and making sure generated text conforms to human values. Unlearning can be used to forget harmful behaviors such as the production of toxic, discriminatory, illegal, or morally undesirable outputs \citep{shevlane2023model,gehman2020realtoxicityprompts,li2024wmdp}, \textit{e.g.}, instructions to build CBRN (chemical, biological, radiological, and nuclear) weapons.   Unlearning, as a safety alignment tool, can happen at the different stages of LLM development, \textit{e.g.}, before, during, or after alignment. Current research has focused on the `pre-alignment' stage \citep{yao2023large}, there may be untapped opportunities in the others.
\textbf{Tab.\,\ref{tab: examples_applications}} exemplifies the response of unlearned LLMs in detoxification \citep{yao2023large} and reducing malicious use of LLMs on the WMDP benchmark \citep{li2024wmdp}. Recall that reject preference optimization-based unlearning \citep{maini2024tofu,jia2024soul} is applied in the detoxification task, which explains why, unlike in WMDP, the post-unlearning generation in detoxification simply responds with ``I'm not able to ...''.


Hallucinations, which involve the generation of false or inaccurate content that may appear plausible, are a significant challenge in LLMs. Previous research has demonstrated that unlearning can reduce LLM hallucinations by targeting and unlearning factually incorrect responses given specific questions \citep{yao2023large}. Since hallucination is likely to be caused by multiple sources, the possible usage is to unlearn factually incorrect data that serve as the source of commonly shared hallucinations or misconceptions.

LLMs are known to generate biased decisions and outputs~\citep{perez2022discovering,tamkin2023evaluating,cui2023holistic}. 
In the vision domain,   unlearning has proven to be an effective tool for reducing discrimination to enable fair decision-making \citep{he2019unlearn,sattigeri2022fair,chen2023fast,dreyer2024hope}. In the language domain, unlearning has been applied to mitigate gender-profession bias \citep{yu2023unlearning} and many other fairness issues \citep{sattigeri2022fair,oesterling2023fair,kadhe2023fairsisa}. However, more opportunities exist, such as unlearning stereotypes in training data.

LLMs are also known to be vulnerable to jailbreaking attacks \citep{wei2023jailbroken,qi2023fine,huang2023catastrophic,zou2023universal} (\textit{i.e.}, adversarially crafted prompts that lead the LLM to generate undesired outputs) as well as poisoning/backdoor attacks~\citep{rando2023universal,carlini2023poisoning,hubinger2024sleeper}. Unlearning can be a natural solution for both types of attacks given the existing success of unlearning as a defense against adversarial attacks in other domains~\citep{wang2019neural,li2021anti,liu2022backdoor,jia2023model}.

\section{Challenges and Overlook}

This work rethinks the paradigm of unlearning for modern LLMs to uncover its under-explored aspects.  
To achieve this, we dissect LLM unlearning into four essential aspects: formulation, methodologies, evaluation metrics, and applications. We show that there are considerable challenges in both foundational research and practical, use case-driven research. These include: 
 {(Generality)} A desired solution for LLM unlearning should take into account the generality of the unlearning target and dataset choice, accommodate various model setups including both white-box and black-box scenarios, and consider the specifics of the unlearning method.
 {(Authenticity)} LLM unlearning should focus on effectively removing both data influence and specific model capabilities, in order to authenticate unlearning across a range of evaluation methods, particularly in adversarial contexts.
 {(Precision)} LLM unlearning should precisely define the scope of unlearning, while ensuring the preservation of general language modeling performance outside this unlearning scope.

{
To advance the above aspects, we point out some future research directions. 

\textit{First}, future research should examine the sensitivity of unlearning methods to factors like optimization hyperparameters, computational resources, and data-model selections. Investigating how unlearning methods transfer and scale across different data-model sizes is key to understanding and enhancing generality.
For example, insights gained from the study of LLM unlearning could catalyze technological advancements in other types of foundation models, \textit{e.g.}, large vision-language models.

\textit{Second}, future work should prioritize robustness and, ideally, unlearning certification. Addressing vulnerabilities to jailbreaking, relearning attacks, and in-context extraction is crucial, as is balancing robustness with LLMs’ emergent capabilities and scalability. Additionally, certified unlearning, inspired by robustness certification in adversarial ML, warrants further exploration. Towards robust unlearning, localization-informed unlearning shows promise with possible dual advantages of efficiency and efficacy.

\textit{Third}, progress in precision requires well-defined method and dataset benchmarks to specify the forget set for unlearning and the evaluation set for generalization testing. For instance, assessing an unlearning method trained on a forget set but evaluated on variations (\textit{e.g.}, watermarked or transformed text) could yield valuable insights into out-of-distribution robustness. Furthermore, new optimization techniques are needed to achieve an optimal balance between unlearning effectiveness and utility preservation. 

\textit{Fourth}, little attention has been given to the ``interpretability'' of LLM unlearning. Interpretability methods, such as saliency maps, example-based explanations, loss landscapes, and training dynamics, offer valuable tools for understanding why unlearning efforts succeed or fail and should be further explored.
}


{
Furthermore, the necessity for regulations or policies to govern unlearning practices is crucial for the future, given the potential implications on privacy, security, and fairness. While existing research has primarily concentrated on auditing unlearning processes related to membership inference, addressing this issue presents an immensely complex challenge. It includes a multitude of factors, including but not limited to data handling/attribution, model governance, transparency, accountability, and verification throughout the unlearning lifecycle. Regulations and policies need to address issues such as data retention, consent management, and the right to be forgotten, which are particularly critical in sensitive domains like healthcare and security applications. Another future effort we encourage is to build an `LLM Unlearning Algorithm Card' that carefully details the involved parties, data aimed to be unlearned, evaluation reports, and the implementation details of the unlearning practice.
}

\section{Acknowledgements}
S. Liu, J. Jia, and Yuguang Yao were supported by the National Science Foundation (NSF) Robust Intelligence (RI) Core Program Award IIS-2207052 and the Cisco Faculty Research Award. 
P. Hase and M. Bansal were supported by NSF-CAREER Award 1846185, NSF-AI Engage Institute DRL-2112635, DARPA MCS Grant N66001-19-2-4031, and Google PhD fellowship.
We also extend our gratitude to the MIT-IBM Watson AI Lab for their support in this project.

\bibliographystyle{icml2024}
\bibliography{refs/MU_ref,reference}

\end{document}